\documentclass[3p,11pt]{elsarticle}

\makeatletter
\def\ps@pprintTitle{%
 \let\@oddhead\@empty
 \let\@evenhead\@empty
 \def\@oddfoot{\centerline{\thepage}}%
 \let\@evenfoot\@oddfoot}
\makeatother

\usepackage{lineno} 
\usepackage{graphicx} 
\usepackage{amssymb} 
\usepackage{array} 
\usepackage[hidelinks]{hyperref}

\journal{a journal}

\biboptions{authoryear}

\begin{document}
\begin{frontmatter}

\title{The relational processing limits of classic and contemporary neural network models of language processing}

\author[mymainaddress,mysecondaryaddress]{Guillermo Puebla \corref{mycorrespondingauthor}}
\address[mymainaddress]{Department of Psychology, School of PPLS, University of Edinburgh, Edinburgh EH8 9JZ, United Kingdom}
\address[mysecondaryaddress]{Universidad de Tarapac\'{a}, General Vel\'{á}squez 1775, Arica, Chile}

\cortext[mycorrespondingauthor]{Corresponding author}
\ead{guillermo.puebla@ed.ac.uk}

\author[thirdaddress]{Andrea E. Martin}
\address[thirdaddress]{Max Planck Institute for Psycholinguistics, Nijmegen, The Netherlands}

\author[mymainaddress]{Leonidas A. A. Doumas}

\begin{abstract}
The ability of neural networks to capture relational knowledge is a matter of long-standing controversy. Recently, some researchers in the PDP side of the debate have argued that (1) classic PDP models can handle relational structure \citep{rogers2008precis, rogers2014parallel} and (2) the success of deep learning approaches to text processing suggests that structured representations are unnecessary to capture the gist of human language \citep{rabovsky2018modelling}. In the present study we tested the Story Gestalt model \citep{john1992story}, a classic PDP model of text comprehension, and a Sequence-to-Sequence with Attention model \citep{Bahdanau2015NeuralMT}, a contemporary deep learning architecture for text processing. Both models were trained to answer questions about stories based on the thematic roles that several concepts played on the stories. In three critical test we varied the statistical structure of new stories while keeping their relational structure constant with respect to the training data. Each model was susceptible to each statistical structure manipulation to a different degree, with their performance failing below chance at least under one manipulation. We argue that the failures of both models are due to the fact that they cannot perform dynamic binding of independent roles and fillers. Ultimately, these results cast doubts on the suitability of traditional neural networks models for explaining phenomena based on relational reasoning, including language processing.
\end{abstract}

\begin{keyword}
relational reasoning \sep generalization \sep language processing \sep neural networks \sep deep learning
\end{keyword}

\end{frontmatter}


\section{Introduction}

The ability to represent and reason in terms of the relations between objects plays a crucial role across many aspects of human cognition, from visual perception \citep{biederman1987recognition}, to higher cognitive processes such as analogy \citep{holyoak2012analogy}, categorization \citep{medin1993respects}, concept learning \citep{doumas2013comparison}, and language \citep{gentner2016language}. Furthermore, comparative evidence suggests that relational thinking may be the key cognitive process distinguishing the abilities of humans from those of other species \citep{penn2008darwin, christie2010hypotheses}. Given the relevance of the capacity to represent and reason about relations across cognitive domains, several computational models in cognitive science have sought to capture its main characteristics and development \citep[e.g.,][]{chen2017generative, falkenhainer1989structure, doumas2008theory, halford1998processing, hummel1997distributed, hummel2003symbolic, kollias2013context, leech2008analogy, lu2012bayesian, Lu4176, van2006neural, yuandomain}.

Computational models of relational thinking differ in their representational assumptions. In the canonical view, relational thinking entails using predicate representations. A predicate is an abstract structure that can be dynamically bound to an argument, specifying a set of properties about that argument \citep{doumas2005approaches}. For example, \textit{predator}(\textbf{x}) specifies a series of properties about the variable \textbf{x} (e.g., carnivore, hunts, etc.). Predicate representations have two main attributes. In the first place, predicates maintain role-filler independence in that at least some aspect of the semantic content of the predicate is invariant with respect to its arguments. For example, \textit{predator}(fox) and \textit{predator}(lynx) will specify the same set of properties (e.g., carnivore, hunts, etc.) about the objects fox and lynx. In the second place, predicates can be dynamically bound to arguments, namely, fillers can be assigned and reassigned to different roles as needed during processing. That predicates can be dynamically bound to arguments allows a given concept to play different roles at different times or in different situations. For example, in a scene where a fox is preying on a hen, but then a lynx comes and eats the fox, the initial binding of fox to \textit{predator} (i.e., predator(fox)) is easily broken and new binding of fox to \textit{prey} (i.e., prey(fox)) is easily formed. Models based on predicates or formally equivalent systems (i.e., systems that perform dynamic binding of independent representations of roles and fillers, or symbolic systems) successfully account for a wide variety of phenomena in the relational thinking literature \citep[for a review see][]{forbus2017representation}.

By contrast, traditional Parallel Distributed Processing (PDP) models explicitly eschew the need for structured representations \citep[see, e.g.,][]{rogers2014parallel}. Representations in a PDP model correspond to patterns of activation across fixed-size layers of units (i.e., an activation vector). These representations are unstructured because relational roles and objects are not independently represented, but instead are represented consecutively, or all of a piece. PDP approaches to relational thinking seek to obtain relational behavior without invoking symbolic machinery \citep{john1992story, john1990learning, kollias2013context, leech2008analogy, yuandomain}. The reasoning behind these models is that if a traditional PDP model successfully performs some relational reasoning task, then predicates are not strictly necessary for that task, and, by extension, might not actually be accurate approximations of human mental representations. Recently, some researchers have argued that PDP models are capable of handling relational knowledge. For example, \citet{rogers2008precis, rogers2014parallel} have proposed that the gestalt models of text comprehension \citep{john1992story, john1990learning, rabovsky2018modelling, rohde2002connectionist} exhibit successful effective role-to-filler binding. Some of this optimism is based, at least partially, on the achievements of deep learning architectures in natural language processing (NLP). For example, \citet{rabovsky2018modelling} argue that the success of Google's neural machine translation (GNMT) system \citep{wu2016google} implies that structured representations are, in fact, an obstacle to accurately capturing the subtle regularities of human language. 

In the present study, we tested the Story Gestalt (SG) model (St. John, 1992) and a Sequence-to-Sequence with Attention (Seq2seq+Attention) model \citep{Bahdanau2015NeuralMT}---the architecture behind the GNMT system---in a series of tasks requiring binding a number of concepts to several roles in a story.  All stories had relational structure in the sense that (1) the thematic roles were organized in specific ways and (2) filling the roles with different concepts yielded different instantiations of the story. In our simulations we trained both models in a large number of these stories to answer questions about the stories and then tested the models with new (untrained) stories. Importantly, we maintained the relational structure of the test stories relative to the training stories while varying their statistical structure (by changing the stories' typical role fillers) in several ways. Next, we describe the generalities of our task and each model's operation in detail.

Our task, based on the original materials of St. John (1992), consists on answering questions about stories generated by a series of (5) scripts. All the scripts describe events as a sequence of propositions where several concepts play different thematic roles: agent-1, agent-2, topic, patient-theme, recipient-destination, location, manner and attribute. As an illustrative example, consider the Restaurant script (see Table 1). This script describes an event where two people go to a restaurant. Each sentence of the Restaurant script defines fillers for some roles. To generate a specific instance of a Restaurant script (i.e., a Restaurant story) the roles are given values corresponding to specific concepts. Table 2 (column 1) presents an example of an instantiated Restaurant story in a pseudo-natural language format. The first sentence of this story corresponds to the proposition: agent-1 = Anne, agent-2 = Gary, topic = decided-to-go, patient-theme = None, recipient-destination = restaurant, location = None, manner = None, attribute = None. Appendix 1 present all possible concepts values for each role. Note that, as illustrated in Table 1, our scripts produce stories with no repeated topic concepts across propositions.

\begin{table}[ht]
\centering
\caption{Restaurant Script.}
\begin{tabular}{ m{10.3cm}}
\hline
\textbf{Script} \\
\hline
1. $<$agent-1$>$ and $<$agent-2$>$ decided to go to restaurant \\
2. Restaurant quality $<$expensive/cheap$>$ \\
3. Distance to restaurant $<$far/near$>$ \\
4. $<$agent-1/agent-2$>$ ordered $<$cheap-wine/expensive-wine$>$ \\
5. $<$agent-1/agent-2$>$ paid bill \\
6. $<$agent-1/agent-2$>$ tipped waiter $<$big/small/not$>$ \\
7. Waiter gave change to $<$agent-1/agent-2$>$ \\
\hline
\textbf{Concept restrictions} \\
\hline
The roles agent-1 and agent-2 are never `Lois' or `Albert' \\
\hline
\textbf{Deterministic rule} \\
\hline
The quality of the restaurant determines the distance completely: $expensive \rightarrow far$, $cheap\rightarrow near$ \\
\hline
\end{tabular}
\label{table:1}
\end{table}

Each script implements a tree structure where each node represents a proposition and each branch of the tree represents a story. The scripts also implement rules that specify the probability of transitioning from one node to another conditioned on the value of a character or location role. For example, a rule in the Restaurant script (see Table \ref{table:1}) specifies that if the restaurant is expensive, it will be located far away.

We trained the models in two different conditions. In the \textit{concept restricted condition}, some character or object names were never used in specific scripts. For example, in the Restaurant stories the characters Lois and Albert were never used to fill the roles agent-1 or agent-2 (see Table \ref{table:1}; \ref{Appendix:A} presents detailed descriptions of the remaining scripts, their concept restrictions and rules). In the \textit{concept unrestricted condition} all concepts were used in all stories. Stories in both conditions were generated according to the following procedure: (1) a script is chosen at random, (2) a sequence of propositions is generated by traversing the probabilistic tree structure of a scrip and (3) character and vehicles names are given specific values (respecting the script's deterministic rule and the script's concept restrictions in the concept restricted condition).

To get a criterion for each model's performance we designed a \textit{baseline test}. In this test we presented the models trained in the unrestricted condition with concept unrestricted stories and asked questions about the stories. The questions corresponded to the concepts filling the topic role. The models generated an answer in the form of a full proposition. The correct answer was the full proposition in which the topic concept was involved. For example, if a proposition in a restaurant story stated that the ``waiter gave change to Anne" and the model was asked about the ``gave'' proposition the correct answer was ``waiter gave change to Anne". Because in our stories there was no repeated topics the correct answer was unequivocal. Table \ref{table:2} presents an example of a Restaurant baseline story, its questions and their corresponding correct answers.

\begin{table}[ht]
\centering
\caption{Example of a Baseline Story (Restaurant).}
\begin{tabular}{ m{6.5cm} m{1.9cm} m{6cm} } 
\hline
\textbf{Story}	& \textbf{Questions} &	\textbf{Criteria} \\
\hline
1. $<$Anne$>$ and $<$Gary$>$ decided to go to restaurant & decided	& $<$Anne$>$ and $<$Gary$>$ decided to go to restaurant \\
2. Restaurant quality $<$expensive$>$ & quality & Restaurant quality $<$expensive$>$ \\
3. Distance to restaurant $<$far$>$ & distance & Distance to restaurant $<$far$>$ \\
4. $<$Anne$>$ ordered $<$cheap-wine$>$ & ordered & $<$Anne$>$ ordered $<$cheap-wine$>$ \\
5. $<$Anne$>$ paid bill & paid	& $<$Anne$>$ paid bill \\
6. $<$Anne$>$ tipped waiter $<$big$>$ & tipped	& $<$Anne$>$ tipped waiter $<$big$>$ \\
7. Waiter gave change to $<$Anne$>$ & gave	& Waiter gave change to $<$Anne$>$ \\
\hline
\end{tabular}
\label{table:2}
\end{table}

\section{Models}

\subsection{Story gestalt model} \label{sg:1}

\begin{figure}[ht]
\centering\includegraphics[width=0.6\linewidth]{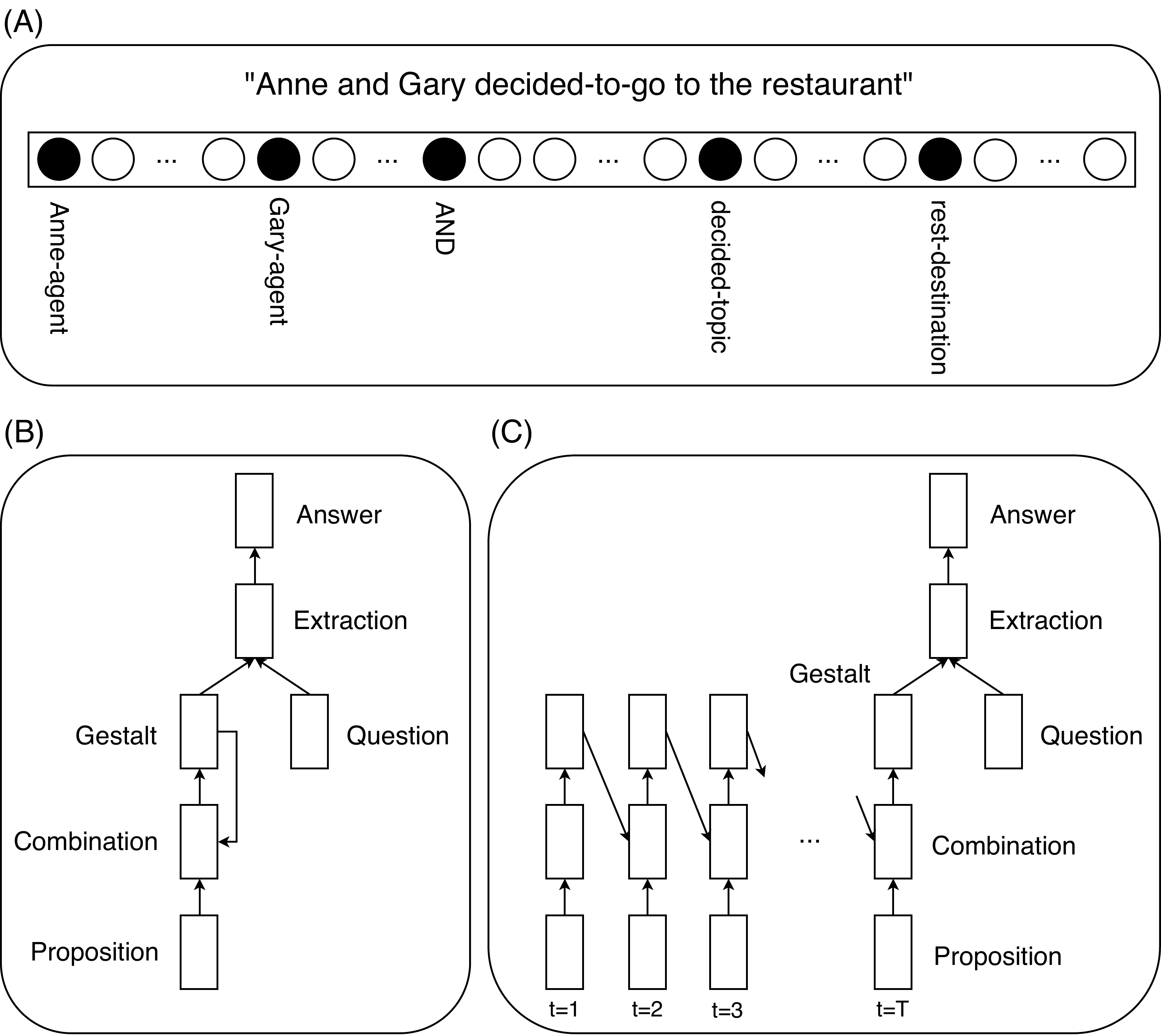}
\caption{Story Gestalt model. (A) An example of an input proposition. (B) Model architecture. (C) Model’s operation unfolded over time. See text for details.}
\label{fig:1}
\end{figure}

The SG model \citep[][see Figure \ref{fig:1}]{john1992story} integrates a sequence of propositions into a distributed representation of a story, which is then used to answer questions about the story. The model represents all propositions in its input layer through 137 localist units coding for each possible filler of each role (e.g., there is a unit coding for Albert-agent and another unit coding for Albert-recipient). To represent a complete proposition, the units coding for the concept filling each role are activated. For example, a representation of the sentence “Anne and Gary decided to go to the restaurant” would consist of a vector of 137 units were the three units coding for Anne-agent, Gary-agent, decided-topic and restaurant-location are set to 1 and all other units are set to 0 (Figure \ref{fig:1}A).

Figure 1B illustrates the SG model's architecture. The model is composed of two subsystems. The first ``comprehension" subsystem (input proposition, combination and gestalt layers), receives each proposition of a story one at the time as input. The activation in the proposition layer feeds forward to the combination and gestalt layers (100 units each). The gestalt layer has recurrent connections to the combination layer, which allows the model to form a representation of the story presented so far (see Figure 1C). The second ``query" subsystem (gestalt, question, extraction and output proposition layers), receives as input the activation of the gestalt layer and the question layer. The question layer (34 units) consists of a vector of units representing all topic concepts in a localist fashion. The extraction layer (100 units) combines the activation of the gestalt and question layers and feeds forward to the output layer, which has the same dimensionality as the input layer.

To train a single story the model is presented with increasing longer sequences of the story propositions and, after each successive sequence, is asked about the last proposition. For example, imagine a story composed by the last three propositions of the Restaurant story in Table \ref{table:2} (i.e., ``Anne paid bill", ``Anne tipped waiter big", ``waiter gave change to Anne"). This story would be trained by presenting the model with the sequences: [``Anne paid bill"], [``Anne paid bill", ``Anne tipped waiter big"] and [``Anne paid bill", ``Anne tipped waiter big'', ``waiter gave change to Anne"]. The question for each sequence would be the topic concept of the last proposition of the sequence (i.e., ``paid", ``tipped" and ``gave") and the target (i.e., what the model was trained to output) would be the last proposition of each sequence (i.e., ``Anne paid bill", ``Anne tipped waiter big", ``waiter gave change to Anne"). The difference between the actual output and the target is used to train the model through a standard gradient descend algorithm. Once trained, the model can recover the full proposition associated with each topic of a story. For example, if a trained SG model is presented with the complete sequence of sentences on Table 2 and then asked about the topic ``decided" (by activating the corresponding localist unit in the question layer) the model would output an activation vector corresponding to the proposition ``Anne and Gary decided to go to the restaurant".

\citet{john1992story} showed that the SG model can recover missing sentences from a story, review its predictions as it encounters new propositions and resolve pronouns. As an example of the model's capabilities consider the case were the model is presented with the complete sequence of propositions on Table \ref{table:2} except for the third (``distance to restaurant far") and is asked about the topic ``distance". In this case the model would output an activation vector corresponding to the proposition ``distance to restaurant far" because in its training data expensive restaurants are always far away (see Table \ref{table:1}).

\subsection{Sequence-to-sequence with attention model} \label{seq2seq:1}

\begin{figure}[ht]
\centering\includegraphics[width=0.8\linewidth]{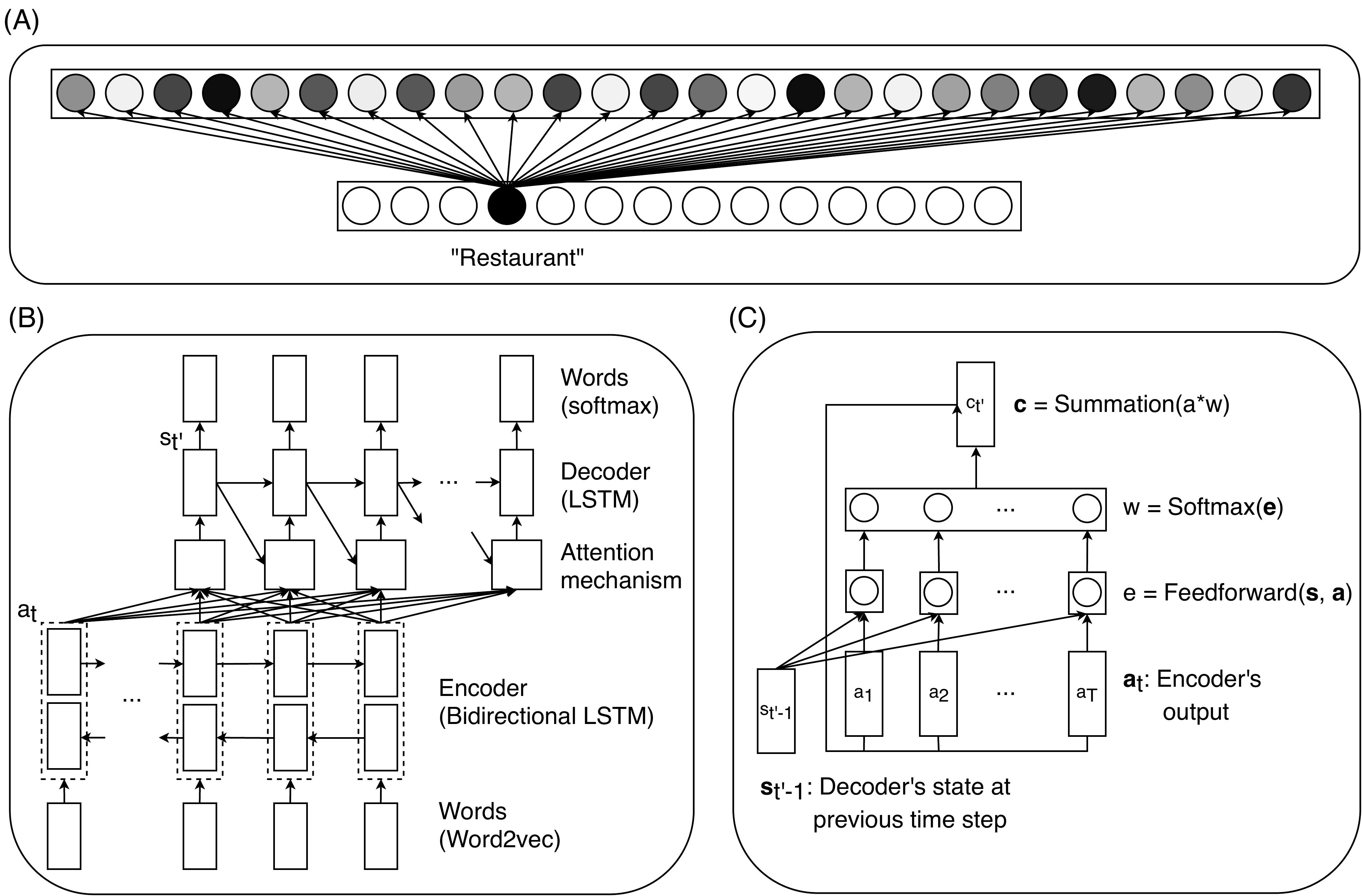}
\caption{Seq2seq model with attention. (A) Input representation. (B) The model’s architecture unfolded over time. (C) The attention mechanism. See text for details.}
\label{fig:2}
\end{figure}

In order to test the performance of a contemporary deep learning system on our task, we implemented a version of the Seq2seq+Attention model \citep[][see Figure \ref{fig:2}]{Bahdanau2015NeuralMT}---a deep neural network architecture designed originally to solve machine translation problems. In translation problems, a source sentence in a given language (e.g., English) has to be translated into a different language (e.g., French). Typically, the source and target sentences have different lengths. In general, a Seq2seq model consist of an encoder network and a decoder network. Both are recurrent neural networks with their own independent time steps (\textit{t} for the encoder and \textit{t'} for the decoder in Figure \ref{fig:2}B). The encoder transforms the input sequence into a sequence of fixed-size vectors and the decoder processes these transformed vectors to get the output sequence. Two important features of our particular implementation are the use of word2vec representations for the input words \citep{mikolov2013distributed} and an attention mechanism that allows the model to selectively “attend” to different parts of the encoder’s output \citep{Bahdanau2015NeuralMT}.

Word2vec embeddings \citep{mikolov2013distributed} are dense distributed representations obtained by extracting the activation vector of the hidden layer of shallow neural network trained to predict the surrounding words given an input word in large corpus of text. Word2vec representations maintain the distributional patterns of similarity between words, such that words used in similar contexts have similar representations \citep[however, see][for evidence of discrepancies between the patterns of similarity between word2vec representations and the patterns of similarity in human word association data]{nematzadeh2017evaluating}. Our version of the Seq2seq+Attention model represents a single word at each time step \textit{t} through a layer with localist units for each unique word in the data set (105 units). To represent a word the corresponding unit is given an activation of 1 while all other units are given an activation of 0 (i.e., a one-hot vector). This one-hot vector is transformed into a word2vec embedding (size 300) by a single feed-forward layer with a fixed set of weights (see Figure \ref{fig:2}A). We did not allow the training process to change these weights.

The encoder (bottom part of Figure \ref{fig:2}B) corresponds to a bidirectional long short-term memory neural network \citep[Bidirectional LSTM,][]{graves2005framewise}. The Bidirectional LSTM is composed of two LSTM neural networks (250 units each in our model). The first LSTM reads the input from the beginning until the end of the sequence while the second reads the sequence in a backwards fashion. At each time step t both LSTMs produce their own output. The full output of the encoder at is the concatenation of the outputs of the forward and backward LSTMs. The encoder's output at each time step \textit{t} can be understood as a summary of all precedent and following words to the current word with an emphasis on the words surrounding it \citep{Bahdanau2015NeuralMT}.

The attention mechanism (center part of Figure 2B and Figure 2C) corresponds to a feed-forward neural network that, at each decoder's time step $t'$, takes as input the decoder previous state $s_{t'-1}$, and all encoder outputs $a_1$ to $a_T$ (see Figure \ref{fig:2}C). This feed-forward network produces a single number $e_t$ for each encoder's output. This number is intended to capture the degree of alignment between the current word in the decoder with each word in the input sequence. This alignment score is normalized using a softmax function, yielding a single attention weight $w_t$ for each encoder's output. The output of the attention mechanism is a context vector $c_t'$, which corresponds to the summation of all encoder's outputs weighted by their corresponding attention weight. In short, the vector $c_t'$ represents a summary of the input words with an emphasis on the words that ``correspond" better with the current output word.

\begin{figure}[ht]
\centering\includegraphics[width=0.8\linewidth]{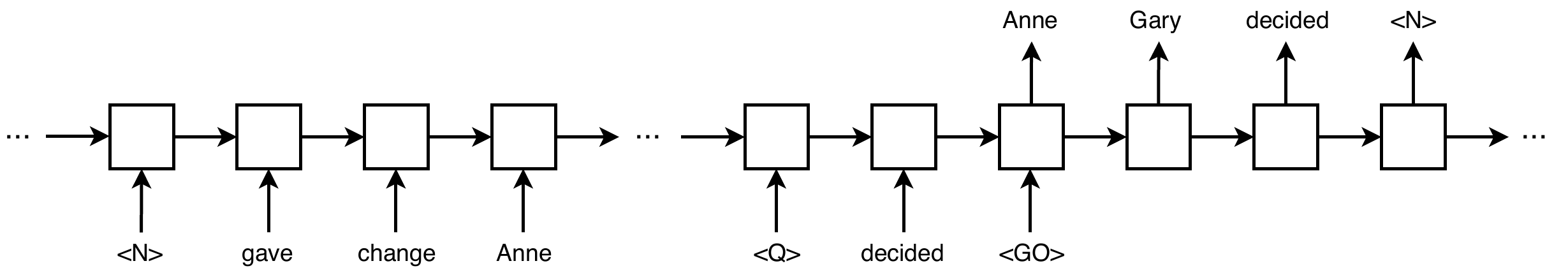}
\caption{Training Process for the Sequence-to-Sequence with Attention Model. See text for details.}
\label{fig:3}
\end{figure}

The decoder (top part of Figure \ref{fig:2}B) corresponds to a standard LSTM network (200 units) followed by feed-forward layer with softmax activation. This layer has a unit for each unique word in the data set (105 units) so that the decoder's output at each time step corresponds to a probability distribution over the dataset vocabulary. The model's answer at each time step is taken to be the word with maximum predicted probability. As this model is designed to receive words as inputs, during training we feed the propositions of our task to the model one word at the time. For each unfilled role we presented the special $<$NONE$>$ word. After presenting the complete story, we input a special word $<$Q$>$ to demarcate the beginning of the question, then we input the topic question, and finally we input a special word $<$GO$>$ to tell the model to start the decoding process. The target output was the sequence of words corresponding to the full proposition involving the topic concept. The difference between the actual output and the target was used to train the model in the same way as in the SG model. Figure \ref{fig:3} presents an example of this process. Here, the Seq2seqs+Attention model (represented by the box) is presented with the complete sequence of words corresponding to the Restaurant story in Table \ref{table:2}. The model is asked about the ``decided" topic and it responds by outputting the sequence of words corresponding to the proposition ``Anne and Gary decided to go to restaurant".

\section{Simulations} \label{task}

We designed three critical tests for the models. Our tests were designed to maintain the relational structure of the stories while varying their statistical properties in comparison to the training data. In short, these tests relied on capturing bindings between roles and fillers in specific instances while ignoring the statistical regularities from the training data. We termed our first test \textit{concept violation}. In this test, we trained the models in the concept restricted condition and then tested them with stories where the agent-1, agent-2 or the patient-theme roles were filled by the restricted concepts. The questions consisted on all the topic concepts of the propositions in which the restricted concepts were used. A role-based answer to the question required using the restricted concept to fill the corresponding role. Table 3 presents an example of a Restaurant concept violation story. In this example, the concepts Albert and Lois had never appeared as agents in any Restaurant story during the model's training. The model was then tested using a story in which Albert or Lois appeared as agents in a Restaurant story by asking, for example, about the ``tipped" proposition. The correct (role-based) answer was ``Lois tipped waiter big". Note that, while the model was trained in stories where Lois appeared as an agent in other locations, and had been trained to output that someone tipped big with other agents, it had never been trained to output the exact proposition ``Lois tipped waiter big". Table \ref{table:3} also presents all the story questions and their corresponding role-based answers.

\begin{table}[ht]
\centering
\caption{Example of a Concept Violation Story (Restaurant). Lois and Albert were restricted from instances of the Restaurant script during training.}
\begin{tabular}{ m{6.5cm} m{1.9cm} m{6cm} } 
\hline
\textbf{Story}	& \textbf{Questions} &	\textbf{Criteria} \\
\hline
1. $<$\textbf{Lois}$>$ and $<$\textbf{Albert}$>$ decided to go to restaurant & decided & $<$\textbf{Lois}$>$ and $<$\textbf{Albert}$>$ decided to go to restaurant \\
2. Restaurant quality $<$expensive$>$ \\		
3. Distance to restaurant $<$far$>$ \\
4. $<$\textbf{Lois}$>$ ordered $<$cheap-wine$>$ & ordered & $<$\textbf{Lois}$>$ ordered $<$cheap-wine$>$ \\
5. $<$\textbf{Lois}$>$ paid bill & paid & $<$\textbf{Lois}$>$ paid bill \\
6. $<$\textbf{Lois}$>$ tipped waiter $<$big$>$ & tipped & $<$\textbf{Lois}$>$ tipped waiter $<$big$>$ \\
7. Waiter gave change to $<$\textbf{Lois}$>$ & gave & Waiter gave change to $<$\textbf{Lois}$>$ \\
\hline
\end{tabular}
\label{table:3}
\end{table}

In our second test, termed \textit{correlation violation}, we presented the models trained in the concept unrestricted condition with stories where we inverted a perfect statistical regularity of the story script. For example, in the Restaurant script the value of the attribute role in the second proposition determines the value of the attribute role in the third proposition in that if the restaurant was cheap it was nearby and if it was expensive it was far away (see Table \ref{table:1}). To create a Restaurant correlation violation story, we switched the value of the attribute role in the third proposition (i.e., a cheap restaurant was now far away, and an expensive restaurant was now nearby). A role-based answer to the questions of this test would use the input concept in the third proposition to fill the attribute role, even though it corresponds to a violation of a correlation seen during training. Table \ref{table:4} presents an example of a Restaurant correlation violation story, its question and corresponding role-based answer. In this example the model had been trained in Restaurant stories where expensive restaurants are always far away and cheap restaurants are always nearby and the model is tested in a Restaurant story where an expensive restaurant is close by. The model is asked about the ``distance" proposition and the correct (role-based) answer is that the restaurant is close by (i.e., ``Distance to restaurant near").

\begin{table}[ht]
\centering
\caption{Example of a Correlation Violation Story (Restaurant).}
\begin{tabular}{ m{6.5cm} m{1.9cm} m{5.3cm} } 
\hline
\textbf{Story}	& \textbf{Questions} &	\textbf{Criteria} \\
\hline
1. $<$Anne$>$ and $<$Gary$>$ decided to go to restaurant\\
2. Restaurant quality $<$\textbf{expensive}$>$\\
3. Distance to restaurant $<$\textbf{near}$>$ & distance & Distance to restaurant $<$\textbf{near}$>$ \\
4. $<$Anne$>$ ordered $<$cheap-wine$>$ \\
5. $<$Anne$>$ paid bill \\
6. $<$Anne$>$ tipped waiter $<$big$>$ \\
7. Waiter gave change to $<$Anne$>$ \\
\hline
\end{tabular}
\label{table:4}
\end{table}

In our third test, termed \textit{shuffled propositions}, we presented the models trained in the concept unrestricted condition with stories where we randomized the order of the propositions. Recall that in our stories there are no repeated topic concepts. As a direct consequence, a role-based answer to a question should use the concepts of the proposition corresponding to each question to fill its roles, ignoring the ordering of the propositions. Table \ref{table:5} presents an example of a Restaurant shuffled propositions story, its questions and their corresponding role-based answers. In this example the model had been trained in stories that followed the same order of propositions as the Restaurant script (see Table \ref{table:1}). The model was presented with sequences of propositions that corresponded to a standard unrestricted Restaurant story, with the only difference being that the order of the propositions was randomized (e.g., the propositions in Table \ref{table:5} are exactly the same as the ones on Table \ref{table:2}), so although the model had received all the individual propositions of the story during training, the model was never trained in the specific sequence being tested. After receiving the propositions, the model was asked about any of the topics of the story. For example, when asked about the ``quality" topic, the correct (role-based) answer was the proposition ``Restaurant quality expensive". It is worth to note that in all our tests the correct (role-based) answers required simply filling the roles of the answer proposition with the concepts that the model had received as input.


\begin{table}[ht]
\centering
\caption{Example of a Shuffled Propositions Story (Restaurant).}
\begin{tabular}{ m{6.5cm} m{1.9cm} m{6cm} } 
\hline
\textbf{Story}	& \textbf{Questions} &	\textbf{Criteria} \\
\hline
\textbf{4.} $<$Anne$>$ ordered $<$cheap-wine$>$ & decided & $<$Anne$>$ and $<$Gary$>$ decided to go to restaurant \\
\textbf{5.} $<$Anne$>$ paid bill & quality & Restaurant quality $<$expensive$>$ \\
\textbf{1.} $<$Anne$>$ and $<$Gary$>$ decided to go to restaurant & distance & Distance to restaurant $<$far$>$ \\
\textbf{3.} Distance to restaurant $<$far$>$ & ordered & $<$Anne$>$ ordered $<$cheap-wine$>$ \\
\textbf{7.} Waiter gave change to $<$Anne$>$ & paid & $<$Anne$>$ paid bill \\
\textbf{6.} $<$Anne$>$ tipped waiter $<$big$>$ & tipped	& $<$Anne$>$ tipped waiter $<$big$>$ \\
\textbf{2.} Restaurant quality $<$expensive$>$ & gave & Waiter gave change to $<$Anne$>$ \\
\hline
\end{tabular}
\label{table:5}
\end{table}

\subsection{Training} \label{training:1}

We trained two versions of the SG model, one in 1,000,000 randomly generated concept restricted stories and another in 1,000,000 randomly generated concept unrestricted stories. We also trained two versions of the Seq2se2+Attention model, one in 500,000 randomly generated concept restricted stories and another in 500,000 randomly generated concept unrestricted stories. We used the Nadam optimization algorithm \citep{dozat2016incorporating} with default learning parameters. All our models were implemented in Keras \citep{chollet2015keras} with TensorFlow backend \citep{abadi2016tensorflow}. Full code for all simulations is available from \url{https://github.com/GuillermoPuebla/RelationReasonNN}.

\subsection{Results} \label{results:1}

\begin{figure}[ht]
\centering\includegraphics[width=0.7\linewidth]{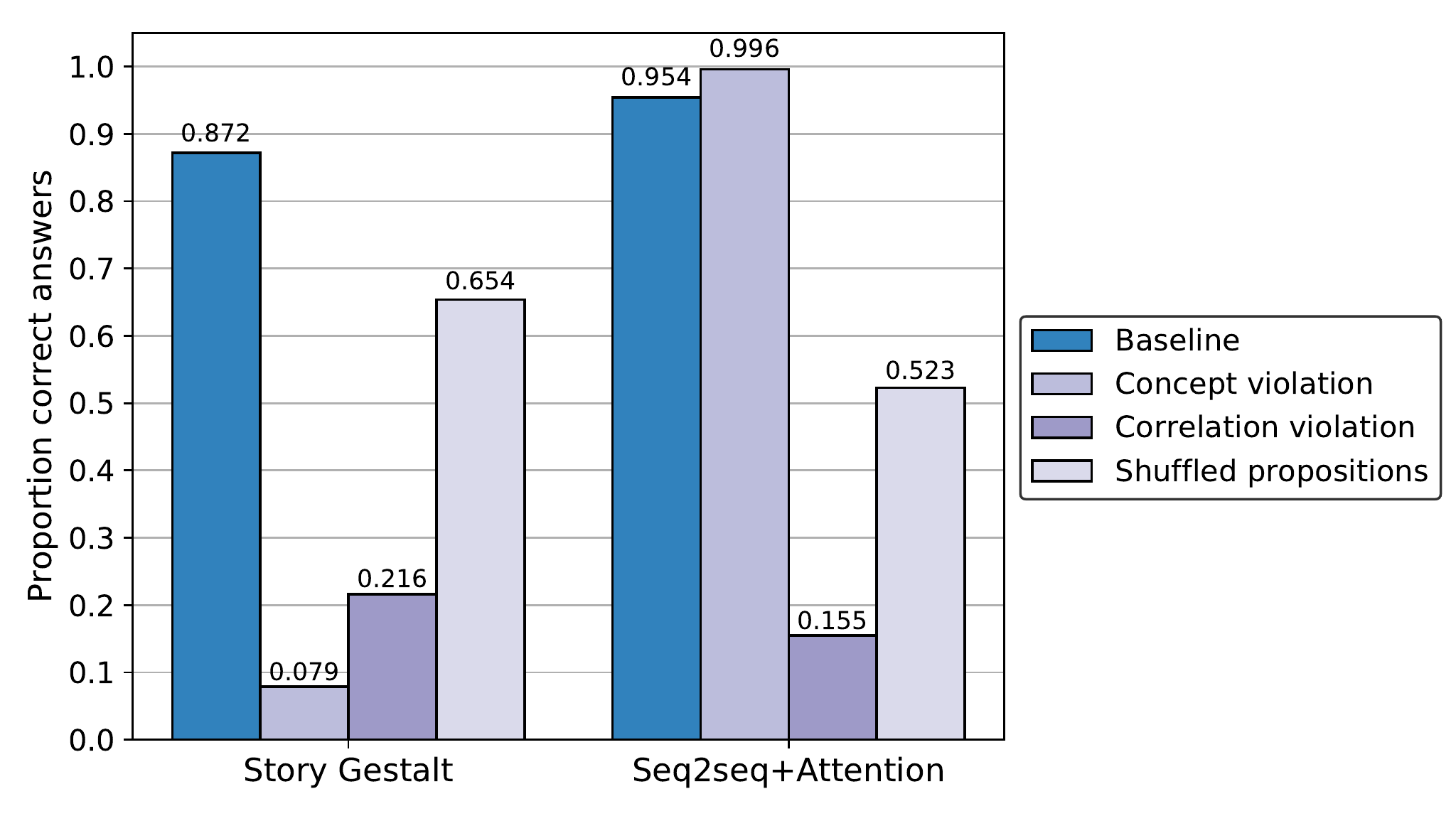}
\caption{Results per test and model.}
\label{fig:4}
\end{figure}

For each of our tests, we created a dataset of stories by generating 1,000,000 stories and saving all unique ones. Due to the combinatorics of concepts and scripts, these datasets had different sizes (baseline and shuffled sentences: 14,652, concept violation: 728, correlation violation: 14,647). For all tests we compared the proposition generated by the model with the role-based answer. We coded the answer as correct (with a value of 1) if the all the concept fillers in the answer corresponded to the concept fillers in the role-based answer and as a non-match (with a value of 0) otherwise. Figure \ref{fig:4} shows the proportion of correct answers per test and model. Recall that in our baseline test we presented the models trained in the concept unrestricted condition with concept unrestricted stories and asked questions about all the propositions in the stories (see Table \ref{table:2} for an example of a Restaurant baseline story, its questions and corresponding correct answers). Because the test stories came from exactly the same distribution as the training stories this test is akin to a recall test of the training dataset. As can be appreciated in Figure \ref{fig:4}, both models performed well in our baseline test. It is noteworthy that the Seq2seq+Attention model showed a better baseline performance than the SG model even though it was trained in half the number of stories.

Recall that in our concept violation test we trained the models in the concept restricted condition and then tested them with stories where the agent-1, agent-2 or the patient-theme roles were filled by the restricted concepts\footnotemark. The questions consisted of all the topics of the propositions in which the restricted concepts were used and a correct (role-based) answer required using the restricted concepts to fill the corresponding roles (see Table \ref{table:3} for an example of a Restaurant concept violation story, its questions and corresponding correct answers). In this test the SG model was unable to use the concepts restricted during training to answer the questions. Instead, the SG model almost invariably filled the roles of the restricted concepts with the most common concepts playing that role during training, which is a direct replication of the results of \citep{john1992story}. For example, if the SG model was presented with a story like the one on Table \ref{table:2} were the roles agent-1 and agent-2 corresponded to the restricted concepts ``Lois" and ``Albert", the model tended to output answers where the agent-1 and agent-2 were any of the other unrestricted agents (e.g., ``Barbara" or ``Clement"). The Seq2seq+Attention model performed significantly better at this test, achieving a slightly better level of accuracy than in the baseline test. The attention mechanism seems to allow this model to apply its word representations to sequences where the words appeared in previously unseen stories.

\footnotetext{Although these concepts were never used the in the context of each specific script, they were seen in the training dataset as a whole. By definition, the output of any traditional neural network to a completely new (unseen) concept depends on its initial weights. Given that these weights are initialized randomly, the behavior of a neural network regarding an unseen input will be essentially random \citep{marcus1998rethinking}.}

Recall that in our correlation violation test we presented the models trained in the concept unrestricted condition with stories where we inverted a perfect statistical regularity of the story script and asked about the proposition that violated the perfect statistical regularity. The correct (role-based) answer required using the input concept even though it violated a statistical correlation from the training dataset. For example, because in the Restaurant script expensive restaurants are always far away, a Restaurant correlation violation story would establish that an expensive restaurant is close by and the correct answer to the ``distance" question would be a proposition that states that the restaurant is indeed close by (see Table \ref{table:4}). Importantly, both models performed poorly in the correlation violation test, in other words, neither model was consistently able to correctly process texts that violated a perfect correlation seen in the training dataset. Such behavior would seem quite unnatural for a human reader as it would amount to, when presented with a proposition stating that a restaurant is close by, answering the question ``where is the restaurant" by stating the restaurant is far away. It is noteworthy that the SG model achieved a higher accuracy than Seq2seq+Attention model in this test (although both models performed quite poorly). We suspect that the same attention mechanism that allows the Seq2seq+Attention model to pass the concept violation test makes it even more likely to overfit to a perfect correlation in the dataset.

Recall that in our shuffled proposition test we presented the models trained in the concept unrestricted condition with concept unrestricted stories where the order of the propositions was randomized. A correct (role-based) answer required to use the concepts of the proposition corresponding to each question to fill its roles, ignoring their ordering (see Table \ref{table:5} for an example). While the randomization of the order of the propositions affected both models, the SG model performed significantly better than the Seq2seq+Attention model in this test. We again hypothesize that the attention mechanism is the main reason for this difference in performance. Unfortunately, because of the length of our stories, taking out the attention mechanism yields the Seq2seq+Attention model unable to pass our baseline test (baseline performance around 0.5), so for now we were not able to test our hypothesis directly.

\section{General discussion}

We tested the relational processing capabilities of the SG model and the Seq2seq+Attention, a classic connectionist model of text comprehension and a contemporary text processing deep learning architecture, respectively. In three critical tests we varied the statistical properties of the test stories while keeping their relational structure intact. Our results show that both models are able to use the statistical regularities of the training data to learn to answer questions correctly for stories that came from the same distribution as the training corpus. More importantly, however, our simulations demonstrate that the performance of both models is severally affected when the statistical properties of the test stories differ from those in the training corpus. Because we kept the relational structure of the test stories intact, our results show clearly that these models are not using the relational information of the stories to answer the questions, but instead they are relying on the statistical regularities of the training dataset. In addition, although the technical advances of Seq2seq+Attention model made it able to pass our concept violation test, this benefit came at the cost of performing worse than the SG model in the correlation violation and shuffled proposition tests.

Both models failed in our tasks demonstrating that they do not perform dynamic binding of independent roles and fillers \citep[see, e.g.,][for discussions]{doumas2005approaches, doumas2012computational}. A model that dynamically binds roles to fillers will easily pass our tests by filling the untrained concepts into the trained roles to answer the questions \citep[see, e.g.,][]{doumas2008theory, falkenhainer1989structure,hummel1997distributed}. It appears that successfully dealing with instances that massively violate the statistics of experience requires representations that explicitly support dynamic binding. 

Our results are highly consistent with the findings of \citet[][see also \cite{Loula2018RearrangingTF}]{Lake2018GeneralizationWS}, who found that sequence-to-sequence models (with and without attention mechanism) failed at a command-to-action translation task that required composing the meaning of new commands formed by using known primitive concepts combined in ways unseen during training. Even in the minority of cases were their models showed behavior that seemed compositional, they did it in a very non-human way (e.g., in one test their best performing model could correctly produce the action sequences corresponding to the instructions ``turn left", and ``jump right and turn left twice", but not the one corresponding to ``jump right and turn left"). 

Truly compositional behavior requires independent representations of objects and roles that can be bound together dynamically (i.e., compositional representations require a solution to the binding problem). In particular, compositionality results when a system can recursively apply predicate representations over other predicate representation \citep[e.g., \textit{loves}(John, \textit{loves}(Mary, Richard)), see][for a discussion]{doumas2005approaches}. We have shown that traditional PDP models (including current deep learning models) do not, as instantiated, perform dynamic binding. As a consequence, these models systematically fail when a task requires violating well learned statistical associations. As such, while there are certainly instances wherein the representations that these models learn will produce the same results as compositional representations, the resulting representations are not truly compositional. To the extent that human mental representations are compositional \citep[see, e.g.,][]{fodor1988connectionism, fodor1975language, lake2017building, marcus2001algebraic, tenenbaum2011grow}, then traditional PDP models fail as successful approximations of human cognition. 

One of the most important evolutionary advantages of relational reasoning is the ability to base inferences on relational roles disregarding the content of their arguments. This capacity allows us to make relational generalizations to completely new inputs \citep{penn2008darwin}. As traditional neural networks can’t, by definition, make use of untrained units to perform successfully in given a task \citep{marcus1998rethinking}, these models rely on spanning the input space to achieve good generalization \citep[see][]{doumas2012computational}. Word embeddings like Word2vec \citep[][cf. \cite{miikkulainen1991natural}]{mikolov2013distributed} can be seen as a technique to deal with this phenomenon. Even though in our Seq2seq+Attention model some concepts were not trained in some contexts, the vector representation of all concepts of a certain type (e.g., agents like ``Anne" and ``Lois") had similar representations because they appear in similar contexts in the Word2vec training dataset. Another strategy to deal with new concepts (or new combinations of concepts) involves directly spanning the input space so that there are no truly new inputs to the model. For example, it is standard practice in neural networks research to make random splits of the data to obtain the training and test datasets. When the data are instantiations of relational structures (as in our tasks) this makes very likely that most objects appear as the fillers of most relational roles in the training dataset, which transforms the relational generalization problem in a interpolation problem, where the correct answer corresponds to an intermediate answer between two known cases \citep[see][for a demonstration of the effects of random versus systematic splits on the training/test datasets]{Lake2018GeneralizationWS}. It is for this reason that traditional PDP models \citep[e.g.,][]{o2002generalizable} and contemporary deep learning models \citep[e.g.,][]{hill2018learning} targeted to solve relational reasoning tasks really on spanning the input space in order to achieve high levels of generalization. Importantly, none of these techniques are solutions to the deeper problem of generalizing to new concepts or new combination of concepts based on abstract relations.

It is, therefore, perhaps unsurprising that deep learning models struggle with tasks that require far transfer \citep[for a review see][]{marcus2018deep}. For example, deep learning models of text comprehension are susceptible to adversarial attacks that add untrained sentences, that share words with the correct answer, to the test text \citep{jia2017adversarial}. Notably, ungrammatical distractor sentences have a stronger adversarial effect than grammatical ones, which suggest that these models are relying in a superficial strategy to solve the reading comprehension task.

However, all of the above is not to say that neural network models cannot, in principle, integrate operations that allow them to implement a truly symbolic dynamic binding system. For example, the symbolic-connectionist models SHRUTI \citep{shastri1993simple}, LISA \citep{hummel1997distributed,hummel2003symbolic}, and DORA \citep{doumas2008theory, doumas2018learning}, use time as a binding signal that allows for role-filler independence and dynamic binding. 

Interestingly, there has been a resurgence of interest on the binding problem in the neural networks \citep{Besold2017NeuralSymbolicLA, franklin2019structured} and computational neuroscience literature \citep{fitz2019neuronal, pina2018oscillations}. Moreover, relational learning and reasoning have become a core topic on deep learning research \citep{Battaglia2018RelationalIB, Greff2015BindingVR, santoro2017simple, hill2018learning} with some deep learning architectures starting to implement operations traditionally associated with symbolic processing such as a content-addressable memory \citep{santoro2016meta, graves2016hybrid, weston2014memory}. Whether these non-traditional neural network architectures are capable of relational reasoning remains an open question that we plan to address in future research. Our results suggest, however, that for a model to successfully account for all aspects of relational processing, it will need to implement a solution to the binding problem. 

Finally, while we herein illustrate the limitations of traditional neural networks when facing relational reasoning tasks, we hope that the results will motivate cognitive scientists and machine learning researchers to tackle the problem of relational learning and reasoning by first tackling the problem of dynamic binding. In the domain of neural network models, doing so will most likely will require us to go beyond the architectural constrains of traditional neural networks.

\section*{Acknowledgements}
The work of Guillermo Puebla was supported by the PhD Scholarship Program of CONICYT, Chile. We thank Hugh Rabagliati for his comments on earlier versions of the manuscript.

\newpage
\bibliography{mybibfile.bib}
\nocite{*}

\newpage
\appendix
\section{Concepts and Story Scripts}
\label{Appendix:A}

\begin{table}[ht]
\centering
\caption{Concepts Used in all the Scripts.}
\begin{tabular}{ m{2.1cm} m{11cm} }
\hline
\textbf{Roles} & \textbf{Concepts}\\
\hline
agents & Albert, Clement, Gary, Adam, Andrew, Lois, Jolene, Anne, Roxanne, Barbara, he, she, jeep, station-wagon, Mercedes, Camaro, policeman, waiter, judge, AND \\
\\
topics & decided, distance, entered, drove, proceeded, gave, parked, swam, surfed, spun, played, weather, returned, mood, found, met, quality, ate, paid, brought, counted, ordered, served, enjoyed, tipped, took, tripped, made, rubbed, ran, tired, won, threw, sky \\
\\
patients or
themes & Albert, Clement, Gary, Adam, Andrew, Lois, Jolene, Anne, Roxanne, Barbara, he, she, jeep, station-wagon, Mercedes, Camaro, ticket, volleyball, restaurant, food, bill, change, chardonnay, prosecco, credit-card, drink, pass, slap, cheek, kiss, lipstick, race, trophy, frisbee \\
\\
recipients or
destinations & Albert, Clement, Gary, Adam, Andrew, Lois, Jolene, Anne, Roxanne, Barbara, he, she, jeep, station-wagon, Mercedes, Camaro, beach, home, airport, gate, restaurant, waiter, park \\
\\
locations & beach, airport, restaurant, bar, race, park \\
\\
manners & long, short, fast, free, pay, big, small, not, politely, obnoxiously \\
\\
attribute & far, near, sunny, happy, raining, sad, cheap, expensive, clear, cloudy \\

\hline
\end{tabular}
\label{table:A6}
\end{table}

\begin{table}[ht]
\centering
\caption{Bar Script.}
\begin{tabular}{ m{10cm} } 
\hline
\textbf{Script} \\
\hline

$<$agent-1$>$ met $<$agent-2$>$ at the bar \\
AND if agent1 = rich (1.0): \\
\hspace{0.5cm}$<$agent-1$>$ enjoyed expensive-wine at the bar \\
AND if agent1 = cheap (1.0): \\
\hspace{0.5cm}$<$agent-1$>$ did not enjoy expensive-wine at the bar \\
$<$agent-2$>$ ordered a drink to the waiter at the bar \\
AND if agent2 = rich (1.0): \\
\hspace{0.5cm}The drink was expensive \\
AND if agent2 = cheap (1.0): \\
\hspace{0.5cm}The drink was cheap \\
OR (2): \\
\hspace{0.5cm}(0.5): \\
\hspace{1cm}$<$agent-2$>$ made a polite pass at $<$agent-1$>$ \\
\hspace{1cm}OR (2): \\
\hspace{1.5cm}(0.3): \\
\hspace{2cm}$<$agent-1$>$ gave a slap to $<$agent-2$>$ \\
\hspace{2cm}$<$agent-2$>$ rubbed cheek \\ 
\hspace{1.5cm}(0.7): \\
\hspace{2cm}$<$agent-1$>$ gave a kiss to $<$agent-2$>$ \\
\hspace{2cm}$<$agent-2$>$ rubbed lipstick \\ 
\hspace{0.5cm}(0.5): \\
\hspace{1cm}$<$agent-2$>$ made a obnoxious pass at $<$agent-1$>$ \\
\hspace{1cm}OR (2): \\
\hspace{1.5cm}(0.7): \\
\hspace{2cm}$<$agent-1$>$ gave a slap to $<$agent-2$>$ \\
\hspace{2cm}$<$agent-2$>$ rubbed cheek \\ 
\hspace{1.5cm}(0.3): \\
\hspace{2cm}$<$agent-1$>$ gave a kiss to $<$agent-2$>$ \\
\hspace{2cm}$<$agent-2$>$ rubbed lipstick \\ 

\hline
\textbf{Concept restrictions} \\
\hline
The roles agent-1 and agent-2 never correspond to `Andrew' or `Barbara' \\
\hline
\textbf{Deterministic rule} \\
\hline
The action of agent-1 determines what agent-2 rubes completely: $slap \rightarrow cheek$, $kiss \rightarrow lipstick$ \\
\hline
\end{tabular}
\label{table:A7}
\end{table}

\begin{table}[ht]
\centering
\caption{Park Script.}
\begin{tabular}{ m{11cm} } 
\hline
\textbf{Script} \\
\hline
$<$agent-1$>$ and $<$agent-2$>$ decided to go to the park \\
The distance to the park was $<$near/far$>$  \\
$<$agent-1$>$ got in $<$vehicle$>$  \\
$<$agent-1$>$ drove $<$vehicle$>$ to the park for a $<$short/long$>$ time  \\ 
$<$agent-1$>$ proceed to the park fast  \\
$<$agent-1$>$ parked at the park for $<$free/pay$>$ \\
The weather was sunny  \\
$<$agent-1$>$ ran through the park  \\
$<$He/She$>$ threw a Frisbee to $<$agent-1/agent-2$>$  \\
\hline
\textbf{Concept restrictions} \\
\hline
The roles agent-1 and agent-2 never correspond to `Clement' or `Roxanne' \\
\hline
\textbf{Deterministic rule} \\
\hline
The distance to the park determines driving time completely: $near \rightarrow short$, $far \rightarrow long$ \\
\hline
\end{tabular}
\label{table:A8}
\end{table}

\begin{table}[ht]
\centering
\caption{Airport Script.}
\begin{tabular}{ m{9cm} } 
\hline
\textbf{Script} \\
\hline
$<$agent-1$>$ decided to go to airport \\
Distance to airport $<$near/far$>$ \\
$<$agent-1$>$ found change \\
$<$agent-1$>$ drove $<$vehicle$>$ to airport $<$short/long$>$ \\
$<$agent-1$>$ ran to gate \\
$<$agent-1$>$ met $<$agent-2$>$ at airport \\
$<$agent-1$>$ $<$agent-2$>$ returned home\\
\hline
\textbf{Concept restrictions} \\
\hline
The roles agent-1 and agent-2 never correspond to `Gary' or `Jolene' \\
\hline
\textbf{Deterministic rule} \\
\hline
The distance to the airport determines driving time completely: $near \rightarrow short$, $far \rightarrow long$ \\
\hline
\end{tabular}
\label{table:A9}
\end{table}

\begin{table}[ht]
\centering
\caption{Beach Script.}
\begin{tabular}{ m{10.5cm} } 
\hline
\textbf{Script} \\
\hline
$<$agent$>$ decided to go to the beach \\
The beach was far away \\
OR (2): \\
\hspace{0.5cm}(0.5):\\
\hspace{1cm}$<$agent$>$ entered $<$vehicle$>$ \\
\hspace{1cm}$<$agent$>$ drove $<$vehicle$>$ to the beach for a long time \\
\hspace{1cm}AND if agent1 = male (1.0): \\
\hspace{1.5cm}$<$agent$>$ proceeded $<$vehicle$>$ to the beach fast \\
\hspace{1.5cm}AND (0.5): \\
\hspace{2cm}The policeman gave a ticket to $<$agent$>$ \\
\hspace{0.5cm}(0.5): \\
\hspace{1cm}$<$agent$>$ drove $<$vehicle$>$ to the beach for a long time \\
AND (0.8): \\
\hspace{0.5cm}$<$agent$>$ swam in the beach \\
\hspace{0.5cm}$<$agent$>$ won the race in the beach \\
\hspace{0.5cm}AND if agent1 = male (0.87): \\
\hspace{1cm}$<$agent$>$ surfed on the beach \\
\hspace{1cm}$<$agent$>$ spun \\
\hspace{0.5cm}AND if agent1 = female (0.33) \\
\hspace{1cm}$<$agent$>$ surfed on the beach \\
AND (0.33): \\
\hspace{0.5cm}$<$agent$>$ played volleyball in the beach \\
OR (2) \\
\hspace{0.5cm}(0.8) \\
\hspace{1cm}The weather was $<$sunny$>$ \\
\hspace{1cm}$<$agent$>$ returned home for a long time \\
\hspace{1cm}$<$agent$>$ was in a $<$happy$>$ mood \\ 
\hspace{0.5cm}(0.2): \\
\hspace{1cm}The weather was $<$cloudy$>$ \\
\hspace{1cm}$<$agent$>$ returned home for a long time \\
\hspace{1cm}$<$agent$>$ was in a $<$sad$>$ mood \\ 
\hline
\textbf{Concept restriction} \\
\hline
The roles recipient and patient never correspond to `Camaro' \\
\hline
\textbf{Deterministic rule} \\
\hline
The weather determines the agent's mood completely: \\
$sunny \rightarrow happy$, $cloudy \rightarrow sad$ \\
\hline
\end{tabular}
\label{table:A10}
\end{table}

\end{document}